\documentclass[a4paper,12pt]{article}
\usepackage[]{graphicx}
\usepackage[]{color}
\makeatletter
\def\maxwidth{ %
 \ifdim\Gin@nat@width>\linewidth
 \linewidth
 \else
 \Gin@nat@width
 \fi
}
\makeatother

\definecolor{fgcolor}{rgb}{0.345, 0.345, 0.345}

\usepackage{framed}
\makeatletter
 {\par\unskip\endMakeFramed%
 \at@end@of@kframe}
\makeatother

\definecolor{shadecolor}{rgb}{.97, .97, .97}
\definecolor{messagecolor}{rgb}{0, 0, 0}
\definecolor{warningcolor}{rgb}{1, 0, 1}
\definecolor{errorcolor}{rgb}{1, 0, 0}
\usepackage{alltt}
\usepackage{varioref}
\usepackage[utf8]{inputenc}
\usepackage[T1]{fontenc}
\usepackage{lmodern}
\usepackage{verbatim}
\usepackage{amsthm}
\usepackage{amsmath}
\usepackage{scrextend} 
\usepackage{wrapfig} 
\usepackage[small]{titlesec} 
\usepackage{enumitem}
\usepackage{longtable} 
\usepackage{amssymb}
\usepackage[super]{nth}
\usepackage[titletoc]{appendix}
\usepackage[export]{adjustbox}
\usepackage{stackengine} 
\usepackage{textcomp} 
\usepackage{footnote} 
\usepackage{scrextend} 
\usepackage[multiple,flushmargin]{footmisc} 
\usepackage{url}
\usepackage[left=1.5cm, right=1.5cm, top=1.5cm, bottom=2cm]{geometry}
\usepackage{subcaption}
\usepackage{textcomp} 
\usepackage{fancyhdr}
\usepackage[numbers]{natbib} 
\usepackage{mhchem} 
\usepackage[pdf]{graphviz}
\usepackage{xcolor}
\usepackage{tikz}
\usepackage{tikz-qtree} 
\usetikzlibrary{trees} 
\usepackage[en-US]{datetime2} 
\DTMlangsetup{ord=raise} 
\usepackage{hyperref} 
\usetikzlibrary{matrix, positioning}
\usetikzlibrary{shapes.misc, positioning}
\usetikzlibrary{arrows}
\usepackage{rotating}
\tikzset{
 treenode/.style = {shape=rectangle, rounded corners,
      draw, align=center,
      top color=white, bottom color=blue!20},
 root/.style  = {treenode, font=\Large, bottom color=red!30},
 env/.style  = {treenode, font=\ttfamily\normalsize},
 dummy/.style = {circle,draw}
}
\usepackage{pgfmath}
\usetikzlibrary{arrows,positioning,fit,shapes,calc}
\usepgflibrary{arrows.meta}
\usetikzlibrary{arrows.meta}
\usetikzlibrary{shapes.geometric}

\usetikzlibrary{arrows.meta}
\usetikzlibrary{decorations.pathreplacing} 
\usepackage{ifthen}
\newcommand{\CplusH}[2]{
 \draw (#1,#2) circle [radius=4.5pt]--++(-3pt,0)node[left]{\quad{\,\large{$C$}}}--++(6pt,0)node[right]{{\large{$H$}}}--++(-3pt,0)--++(0,0pt)--++(0,-3pt)--++(0,6pt);}
 \newcommand{\XplusH}[2]{
 \draw (#1,#2) circle [radius=4.5pt]--++(-3pt,0)node[left]{\quad{\,\large{$X$}}}--++(6pt,0)node[right]{{\large{$H$}}}--++(-3pt,0)--++(0,0pt)--++(0,-3pt)--++(0,6pt);}
 
\newcommand{\cph}{$C$~\stackinset{c}{0pt}{0pt}{-0.2pt}{\normalsize{\textbigcircle}}{
\stackinset{c}{-1.8pt}{0pt}{1.5pt}{\tiny{+}}}~$H$}
\newcommand{\cphu}{$C$~$\mathrel{\raisebox{0pt}{\textnormal{\small{\textbigcircle}}}}$\!\!\!$\mathrel{\raisebox{2.2pt}{\textnormal{\tiny{+}}}}$ $H$}

\DeclareMathOperator{\NFP}{NFP}

\usepackage{arydshln}
\usepackage{etoolbox}
\newtoggle{originalversion}

\togglefalse{originalversion} 
\newrobustcmd{\org}[1]{\iftoggle{originalversion}{\sout{#1}}{}}
\newtoggle{correctversion}
\toggletrue{correctversion} 
\newrobustcmd{\cor}[1]{\iftoggle{correctversion}{#1}{}}
\newrobustcmd{\oc}[2]{\org{#1}\cor{\color{red}#2\color{black}}}
\renewcommand{\paragraph}[1]{\medskip\par\noindent\textbf{#1}\newline}

\newcommand{\HRule}{\rule{\linewidth}{0.5mm}}

\begin{document}
\textbf{ }
\HRule
\\
\centering
{\large \bfseries Synergy Effect
\\
between Convolutional Neural Networks
\\
and the Multiplicity of SMILES
\\
for Improvement of Molecular Prediction}
\HRule
\\
\vspace{0.5cm}
\textbf{Talia B. Kimber\footnote{University of Geneva, Research Center for Statistics, 1211 Geneva, Switzerland\label{refnote1}}},
\textbf{Sebastian Engelke\footref{refnote1}},
\textbf{Igor V. Tetko\footnote{BIGCHEM GmbH and Helmholtz Zentrum Muenchen - German Research Center for Environmental Health (GmbH), Ingolstaedter Landstrasse 1, D-85764 Neuherberg, Germany}}
\\
\vspace{0.3cm}
\textbf{Eric Bruno\footref{refnote2}},
\textbf{Guillaume Godin\footnote{Firmenich SA, Corporate R\&D Division, 1211 Geneva, Switzerland\label{refnote2}}$^{,}$\footnote{Corresponding author. Email address: guillaume.godin@firmenich.com}}
\vspace{0.8cm}
\\
\abstract{
In our study, we demonstrate the synergy effect between convolutional neural networks and the multiplicity of SMILES. The model we propose, the so-called Convolutional Neural Fingerprint (CNF) model, reaches the accuracy of traditional descriptors such as Dragon (\citet{mauri06}), RDKit (\citet{rdkit}), CDK2 (\citet{willighagen17}) and PyDescriptor (\citet{masand17}). Moreover the CNF model generally performs better than highly fine-tuned traditional descriptors, especially on small data sets, which is of great interest for the chemical field where data sets are generally small due to experimental costs, the availability of molecules or accessibility to private databases. We evaluate the CNF model along with SMILES augmentation during both training and testing. To the best of our knowledge, this is the first time that such a methodology is presented. We show that using the multiplicity of SMILES during training acts as a regulariser and therefore avoids overfitting and can be seen as ensemble learning when considered for testing.}
\section{Introduction}
Predicting molecular activities using machine learning algorithms such as deep neural networks is of high interest for chemical or pharmaceutical companies. Recent publications have shown the capacity of statistical learning to efficiently model chemical and biological phenomena at a molecular level with high accuracy (\citet{baldi18}).
\par One of the great challenges in chemoinformatics is to model a chemical structure. Indeed the physical and chemical aspects of these structures make for highly complex objects.
\par A chemical structure is often represented through the use of a connected and undirected graph, called molecular graph, which corresponds to the structural formula of the chemical compound. Vertices of the graph correspond to atoms of the molecule and edges to chemical bonds between these atoms. 
\par From a computational point of view, a molecular graph remains a complex object and finding a numerical representation of a chemical structure is necessary in the context of machine learning. Consequently the concept of a fingerprint was developed. A fingerprint is a numerical representation which encodes molecular information (\citet{todeschini09}). One of the reasons for developing fingerprints is to extract substructures from molecules. It is believed that molecules with similar groups of substructures will have similar properties (\citet{muegge16}).
\par Another notation derived from the molecular graph is the Simplified Molecular Input Line Entry System (SMILES) suggested by \citet{Weininger88}. 	SMILES is a linearisation of a given molecular graph, in the form of a single line text, obtained by enumerating nodes and edges following a certain topological ordering. Although SMILES encodes information on the local structure of the graph, a problem arises when linearising a graph. Cycle, branch and stereochemistry breaks can create ambiguities, especially in the training of a neural network.
\par A number of equally valid SMILES strings can be obtained for a single molecule since several topological orderings may exist for the corresponding graph. Indeed starting the enumeration from different nodes and following different paths of a molecular graph will lead to different SMILES representations of the same molecule. The number of possible SMILES grows with the complexity of the graph.
\par In order to have a one-to-one relationship between a molecule and its SMILES representation, some algorithms have been developed to choose one SMILES, called the canonical SMILES. There is no universal canonical SMILES since each toolkit has its own canonisation algorithm. Similar molecules do not necessarily have similar canonical SMILES representations. 
\par The compact notation of SMILES allows for efficient numerical computation and is therefore used in this paper to predict molecular properties.
\par Widely used techniques to extract patterns in sequence modelling, such as SMILES, are Recurrent Neural Networks (RNN) and Convolutional Neural Network (CNN) (\citet{bai18}). To account for ambiguities of cycle, branch and stereochemistry breaks, we suggest providing multiple versions of SMILES, which should be able to expose various views of the molecular graph and allow for a more global understanding of the structure of the chemical compound.
\par In this paper we expose a modelling approach of chemical structures inspired by the ResNet model by \citet{ren15} which uses CNNs that combine both linear and hierarchical layers. Our contribution lies in the combination of CNN as a feature extraction method and the multiplicity of SMILES as a means to achieve data augmentation as suggested by \citet{bjerrum17}.
\section{Related work}
In chemoinformatics, the use of SMILES as input for machine learning algorithms is not rare (\citet{jastrzebski16}, \citet{olier17}, \citet{lee18}) and applying RNNs to SMILES has become more and more common. For example, \citet{segler17} train RNNs as generative models for molecules and the deep RNN SMILES2vec model proposed by \citet{goh17smiles2vec} is used to predict chemical properties.
\par In the framework of SMILES, \citet{bjerrum17} suggests that increasing the size of a dataset by considering several SMILES for one molecule is a way to achieve data augmentation. Previous publications have shown the effectiveness of data augmentation in deep learning applied to several different fields such as image classification (\citet{DA_Perez}), time series classification (\citet{fawaz18}), face recognition (\citet{Lv17}), relation classification (\citet{DA_DNN}) and soft biometrics classification (\citet{DA_CNN}). In this context, \citet{goh17} took advantage of the multiplicity of SMILES in RNN models.
\par Although SMILES have been additionally processed through RNN architectures, we propose here to use CNNs, following \citet{bai18} who have demonstrated the advantage of convolutional operators in sequence processing.
\par As an alternative to SMILES, recent research was conducted to define fingerprints directly from 2D molecular graphs, leading to neural fingerprints and the ConvGraph model, as proposed by \citet{duvenaud15}. The idea behind neural fingerprints is to apply convolutions in order to obtain feature extraction. CNNs applied to string vectors have shown their effectiveness in different fields such as sentence classification, with the TextCNN model introduced by \citet{kim14}.
\par Considering all of the above, we build a CNN model which acts as a feature extractor, leading to a neural fingerprint. We not only use SMILES as inputs but also take advantage of the multiplicity of SMILES as a means of obtaining data augmentation to cope with cycle, branch and stereochemistry breaks.
\section{Convolutional Neural Fingerprint}
In this section, we explain the Convolutional Neural Fingerprint (CNF) model, which heavily relies on the convolutional graph introduced by \citet{duvenaud15}, CNN for sentence classification (\citet{kim14}) and the architecture of the ResNet model suggested by \citet{ren15}.
\par The key ideas behind the convolutional graph are the hierarchical convolutional layers and a matrix multiplication which acts as a hash function. In the original paper by \citet{duvenaud15}, the convolutional graph operates directly on a molecular graph, which is known to be computationally inefficient. This is why we suggest implementing a CNN on the SMILES representation of a molecule. As a matter of fact, convolutional layers applied to string vectors have shown their effectiveness in sentence classification (\citet{kim14}) and SMILES are precisely single line characters which encode the information of a molecular graph. 
\par The CNF model applies to the one-hot encoding of SMILES convolutional layers followed by matrix multiplications, which we denote  \cph, where $H$ is an embedding matrix. The idea behind applying convolutional layers is to obtain neighbouring atoms in a chemical structure and \oc{the}{} multiplication can be interpreted as an embedding on a latent space. This  \cph\, implementation is highly linked to the Morgan fingerprint (\citet{roger10}), where on the one hand, relevant substructures, or neighbourhood of atoms, are encoded and on the other hand, features are hashed.
\par
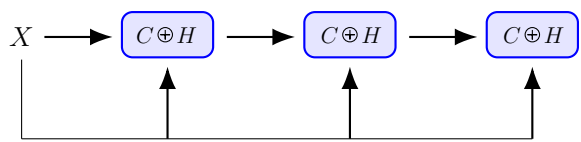
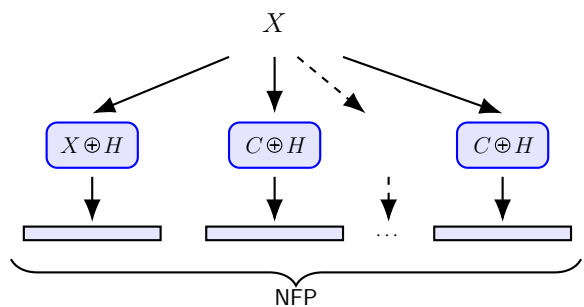
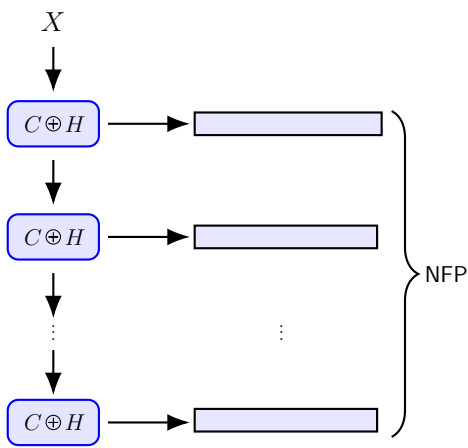
\begin{figure}
    \centering
    \begin{subfigure}{1\textwidth}
    \centering
\begin{tikzpicture}[scale=0.6, transform shape,
longblock/.style={draw=black, thick, fill=blue!10, 
text width=10em, text centered, 
minimum height=2em] }, 
block/.style ={rectangle, draw=blue, thick, fill=blue!20, 
text width=5em, text centered, rounded corners, 
minimum height=4em, fill opacity=0.5},
arrowvert/.style={-{Latex[length=3mm]},thick }
 ]
\draw(1,13.25)node[]{\textsf{\Large{$X$}}};
\draw[arrowvert](1.5,13.25) -- (3,13.25);
\draw [block](3.2,12.8)+(0,0) rectangle ++(2,1); 
\CplusH{4.15}{13.3}
\draw[arrowvert](5.5,13.25) -- (7,13.25);
\draw [block](7.2,12.8)+(0,0) rectangle ++(2,1); 
\CplusH{8.2}{13.3}
\draw[arrowvert](9.5,13.25) -- (11,13.25);
\draw [block](11.2,12.8)+(0,0) rectangle ++(2,1); 
\CplusH{12.2}{13.3}
\draw(1,12.75) -- (1,11);
\draw(1,11) -- (12.2,11);
\draw[arrowvert](4.2,11) -- (4.2,12.6);
\draw[arrowvert](8.2,11) -- (8.2,12.6);
\draw[arrowvert](12.2,11) -- (12.2,12.6);
\end{tikzpicture}
	\caption{Diagram of the CNF model following the ResNet architecture. The input $X$ is the one-hot encoding of a SMILES. \cphu denotes a convolution plus a hashing layer.} 
	\label{fig:res_net}
    \end{subfigure}
    \\
    \vspace{2cm}
    \begin{subfigure}{1\textwidth} 
      \centering 
\begin{tikzpicture}[scale=0.6, transform shape,
longblock/.style={draw=black, thick, fill=blue!10, 
text width=10em, text centered, 
minimum height=2em] }, 
block/.style ={rectangle, draw=blue, thick, fill=blue!20, 
text width=5em, text centered, rounded corners, 
minimum height=4em, fill opacity=0.5},
arrowvert/.style={-{Latex[length=3mm]},thick }
 ]
\draw[longblock] (0,4.5) rectangle (3,4.2);
\draw[longblock](4,4.5) rectangle (7,4.2);
\draw[] (8,4.3)node{$\cdots$};
\draw[longblock] (9,4.5) rectangle (12,4.2);
\draw [block](0.5,5.8)+(0,0) rectangle ++(2,1); 
\XplusH{1.5}{6.3}
\draw[block] (4.5,5.8)+(0,0) rectangle ++(2,1); 
\CplusH{5.5}{6.3}
\draw [block](9.5,5.8)+(0,0) rectangle ++(2,1); 
\CplusH{10.5}{6.3}
\draw[arrowvert](1.5,5.6)--(1.5,4.6);
\draw[arrowvert](5.5,5.6)--(5.5,4.6);
\draw[arrowvert](10.5,5.6)--(10.5,4.6);
\draw[arrowvert,dashed](8,5.6)--(8,4.6);
\draw(5.5,9)node{\textsf{\Large{$X$}}};
\draw[arrowvert](4.5,8.25)--(1.5,7);
\draw[arrowvert](5.5,8.25)--(5.5,7);
\draw[arrowvert,dashed](6.,8.25)--(7.5,7);
\draw[arrowvert](7.,8.25)--(10.5,7);
\draw [decorate,decoration={brace,amplitude=10pt,mirror,raise=0pt},xshift=-8pt,yshift=8pt, thick]
(0.,3.5) -- (12.5,3.5) node [black,midway,below,yshift=-0.5cm] {\large{\textsf{NFP}}};
\end{tikzpicture} 
	\caption{Diagram of the CNF model with flat convolutional layers. The input $X$ is the one-hot encoding of a SMILES. The first layer hashes $X$ without performing a convolution. The following layers use convolution, plus hashing, denoted \cphu. After each layer, we sum the columns of the matrix, namely a pooling layer. Concatenating all these vectors leads to the neural fingerprint.} 
	\label{fig:flat}
    \end{subfigure}
\\
\vspace{2cm}    
    \begin{subfigure}{1\textwidth} 
      \centering 
\begin{tikzpicture}[scale=0.6, transform shape,
longblock/.style={draw=black, thick, fill=blue!10, 
text width=10em, text centered, 
minimum height=2em] }, 
block/.style ={rectangle, draw=blue, thick, fill=blue!20, 
text width=5em, text centered, rounded corners, 
minimum height=4em, fill opacity=0.5},
arrowvert/.style={-{Latex[length=3mm]},thick }
 ]
\draw(1,13.25)node[]{\textsf{\Large{$X$}}};
\draw[arrowvert](1,12.7) -- (1,11.7);
\draw [block](0.0,10.5)+(0,0) rectangle ++(2,1); 
\CplusH{1}{11}
\draw [block](0.0,8)+(0,0) rectangle ++(2,1); 
\CplusH{1}{8.5}
\draw [block](0.0,3.9)+(0,0) rectangle ++(2,1); 
\CplusH{1}{4.4}
\draw(1,6.5) node{$\vdots$}; 
\draw[arrowvert](1,10.2) -- (1,9.2);
\draw[arrowvert](1,7.7) -- (1,6.7);
\draw[arrowvert](1,6) -- (1,5);
\draw(6,6.5) node{$\vdots$}; 
\draw [longblock](4.1, 10.75)+(0,0) rectangle ++(4.1,0.5); 
\draw [longblock](4.1,8.25)+(0,0) rectangle ++(4,0.5); 
\draw [longblock](4.1,4.2)+(0,0) rectangle ++(4,0.5); 
\draw[arrowvert](2.2,4.4) -- (4,4.4);
\draw[arrowvert](2.2,8.5) -- (4,8.5);
\draw[arrowvert](2.2,11) -- (4,11);
\draw [decorate,decoration={brace,amplitude=10pt,mirror,raise=0pt},xshift=-8pt,yshift=8pt, thick]
(8.7,3.8) -- (8.7,11) node [black,midway,xshift=1.2cm] {\large{\textsf{NFP}}};
\end{tikzpicture}
	\caption{Diagram of the CNF model with hierarchical convolutional layers. The input $X$ is the one-hot encoding of a SMILES. We apply a convolutional plus a hashing layer to $X$, denoted \cphu. We obtain a first vector by summing the columns of the matrix, namely a pooling layer. The result of \cphu is used as an input of the next layer, from which we obtain a second vector after pooling. Concatenating all these vectors leads to the neural fingerprint.} 
	\label{fig:hier}
    \end{subfigure}
    \caption{Diagram of the CNF model.} 
    \label{fig:CNFarchi}
\end{figure}

 The architecture of the CNF model resembles the one of the ResNet model (\citet{ren15}). Indeed both hierarchical and flat convolutional layers are used (see Figure~\ref{fig:res_net}).
\par The neural fingerprint is obtained as follows: a pooling layer is applied to each \cph \, operation, leading to a vector. For the CNF model, the pooling layer corresponds to summing the rows of the matrix \cph. Both the hashing $H$ and the pooling make for a nonlinear model. The neural fingerprint is obtained by concatenating all the vectors obtained after pooling. Figures~\ref{fig:flat} and \ref{fig:hier} show the architecture of the CNF model in the case where the layers are strictly flat and strictly hierarchical, respectively.
\par An advantage of the CNF model is that it maps a sparse matrix, the one-hot encoding of a SMILES, into a dense vector, the neural fingerprint. The latter can be plugged directly into standard machine learning algorithms such as neural network, support vector machine and random forest. \citet{duvenaud15} have shown that similar molecules have similar (in the Euclidian sense) neural fingerprints, even when the model has not been trained. In the CNF case, two SMILES with similar spelling will be close, in the Euclidian sense, in the latent space. This observation was already underlined in sentence classification when dealing with ``semantically close words'' (\citet{kim14}). In Section~\ref{supp_mat}, we show the evolution of the distance matrix of the neural fingerprints as the CNF model trains (see Figure~\ref{dist_mat_close_up}).
\section{Data Augmentation}\label{DA}
One of the main objectives of this research is to determine whether the multiplicity of SMILES can be used as means of providing data augmentation and therefore help the learning process of a neural network. The key idea behind data augmentation is to augment the number of observations from a given data set to obtain better performances when training machine learning algorithms.
\par One of the major risks when fitting a model is overfitting the data and this risk becomes greater when dealing with complex non-linear models, such as neural networks (\citet{bishop1995}, \citet{tetko95}). Regularisation is a common way to prevent overfitting (\citet{murphy2012}) and data augmentation is an effective technique of regularisation (\citet{DA_Reg}).
\par In chemoinformatics, \citet{bjerrum17} puts forward the idea of SMILES enumeration as a method of data augmentation. As mentioned previously, one molecule can have various SMILES representations. This multiplicity depends both on the atom where the enumeration starts and the path followed along the 2D graph of the molecular structure. Data are therefore created by exploring different directions of a given graph. Figure~\ref{fig:3SMILES} shows the multiplicity of SMILES of the ethylcyclopropane molecule. We underline the fact that the two first SMILES have the same starting point, but the paths along the 2D graph are different and therefore lead to different SMILES.
\par The intuitive reason for believing that SMILES augmentation improves predictive performances is the following: a molecular graph is a cyclic graph, which contains all the structural information of a chemical compound, as a whole. However when constructing an associated SMILES, we choose to enumerate recursively the atoms present in the molecule in a certain sequence, which is equivalent to linearising a graph. Nevertheless information on the cyclic properties of the molecule is buried when using this line notation; a character indicating the beginning of a cycle or a branch may be separated from its closing character by many other strings, making the understanding of cycles difficult for a neural network. In this sense, SMILES are an acyclic graph of a molecule. Generating several valid SMILES, each with a different starting point and following different paths, will lead to different ``spellings'' of the same molecule, exposing many angles of the same item. By enumerating SMILES, we transfer the perspective from a local to a more global one. Similar to image classification, multiple SMILES allow for several views of the same object.
\par In our simulations, we make the distinction between the augmentation of SMILES during training and the augmentation during inference. Indeed both \citet{bjerrum17} and \citet{lee18} have shown the effectiveness of SMILES augmentation during training. However, to our knowledge, no simulations have been performed using SMILES augmentation during both training and testing.
\par SMILES augmentation during testing can be regarded as ensemble learning. When provided with one molecule, we generate several valid SMILES along with their respective model outputs. The aggregation of the outputs allows for a consensus effect. In the regression case, the outputs are averaged whereas in the classification case, the classification is done based on the majority vote of the different outputs.
\par The results of the simulations in the next section show that the technique of SMILES augmentation during training and testing improves predictive performances. 
\newcommand{\shiftx}{-5}
\newcommand{\shifty}{0}
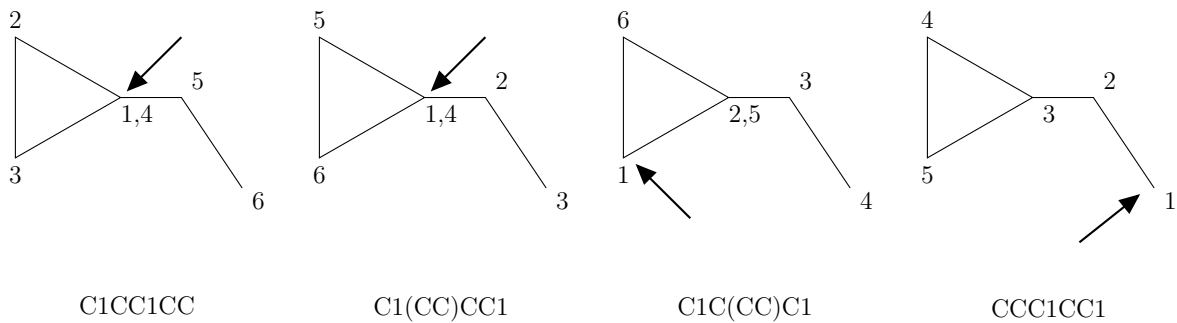
\begin{figure}
\centering
\begin{tikzpicture}[scale=0.8,transform shape,>=triangle 45]

\renewcommand{\shiftx}{-5}
\renewcommand{\shifty}{0}
\draw(++0+\shiftx,++0+\shifty)-- (++0+\shiftx,++1+\shifty)--(++0+\shiftx,++1+\shifty)--(++1.73+\shiftx,++0+\shifty)--(++0+\shiftx,++-1+\shifty)--(++0+\shiftx,++0+\shifty);
\draw(1.73+\shiftx,0+\shifty)--(2.73+\shiftx,0+\shifty)--(3.73+\shiftx,-1.5+\shifty);
\draw[<-, thick](1.83-5,0.1+\shifty)--(2.73-5,1+\shifty);
\draw++(2+\shiftx,0+\shifty)node[below]{1,4};
\draw++(4+\shiftx,-2+\shifty)node[above]{6};
\draw++(3+\shiftx,0+\shifty)node[above]{5};
\draw++(0+\shiftx,1+\shifty)node[above]{2};
\draw++(0+\shiftx,-1+\shifty)node[below]{3};
\draw(-3+0,++-3.45)node{C1CC1CC}; 

\renewcommand{\shiftx}{0}
\renewcommand{\shifty}{0}
\draw(0+\shiftx,0+\shifty)-- (0+\shiftx,1+\shifty)--(0+\shiftx,1+\shifty)--(1.73+\shiftx,0+\shifty)--(0+\shiftx,-1+\shifty)--(0+\shiftx,0+\shifty);
\draw(1.73+\shiftx,0+\shifty)--(2.73+\shiftx,0+\shifty)--(3.73+\shiftx,-1.5+\shifty);
\draw[<-, thick](1.83+\shiftx,0.1+0)--(2.73+\shiftx,1+0);

\draw++(2+\shiftx,0+\shifty)node[below]{1,4};
\draw++(4+\shiftx,-2+\shifty)node[above]{3};
\draw++(3+\shiftx,0+\shifty)node[above]{2};
\draw++(0+\shiftx,1+\shifty)node[above]{5};
\draw++(0+\shiftx,-1+\shifty)node[below]{6};
\draw(2+0,++-3.5)node{C1(CC)CC1}; 

\renewcommand{\shiftx}{5}
\renewcommand{\shifty}{0}
\draw(0+\shiftx,0+\shifty)-- (0+\shiftx,1+\shifty)--(0+\shiftx,1+\shifty)--(1.73+\shiftx,0+\shifty)--(0+\shiftx,-1+\shifty)--(0+\shiftx,0+\shifty);
\draw(1.73+\shiftx,0+\shifty)--(2.73+\shiftx,0+\shifty)--(3.73+\shiftx,-1.5+\shifty);
\draw[<-, thick](+0.2+5,-1.1+0)--(1.1+5,-2+0);
\draw++(2+\shiftx,0+\shifty)node[below]{2,5};
\draw++(4+\shiftx,-2+\shifty)node[above]{4};
\draw++(3+\shiftx,0+\shifty)node[above]{3};
\draw++(0+\shiftx,1+\shifty)node[above]{6};
\draw++(0+\shiftx,-1+\shifty)node[below]{1};

\draw(7+0,++-3.5)node{C1C(CC)C1};

\renewcommand{\shiftx}{10}
\renewcommand{\shifty}{0}
\draw(++0+\shiftx,++0+\shifty)-- (++0+\shiftx,++1+\shifty)--(++0+\shiftx,++1+\shifty)--(++1.73+\shiftx,++0+\shifty)--(++0+\shiftx,++-1+\shifty)--(++0+\shiftx,++0+\shifty);
\draw(1.73+\shiftx,0+\shifty)--(2.73+\shiftx,0+\shifty)--(3.73+\shiftx,-1.5+\shifty);
\draw[<-, thick](3.5+10,-1.6+0)--(2.5+10,+-2.4);
\draw++(2+\shiftx,0+\shifty)node[below]{3};
\draw++(4+\shiftx,-2+\shifty)node[above]{1};
\draw++(3+\shiftx,0+\shifty)node[above]{2};
\draw++(0+\shiftx,1+\shifty)node[above]{4};
\draw++(0+\shiftx,-1+\shifty)node[below]{5};

\draw(12+0,++-3.5)node{CCC1CC1};

\end{tikzpicture}
	\caption{Consider the molecule ethylcyclopropane given by the chemical formula \ce{C5H10}. The figure illustrates the 2D and SMILES representation of the molecule. Four diagrams are displayed. In each diagram, the arrow points at the atom where the enumeration starts for the SMILES construction. The path along the 2D graph can be followed with the numbers starting at 1 next to the nodes. All four SMILES are equally valid for this molecular structure.} 
	\label{fig:3SMILES}
\end{figure}
%
\section{Results}
In this section we discuss the results of our simulations, which are performed on OCHEM (\citet{Sushko2011}), an online chemical modelling platform, which itself uses the Python library DeepChem (\citet{deepchem16}) and the open source chemoinformatics software RDKit (\citet{rdkit}). We stress here the fact that a new RDKit feature has been developed to generate all possible valid SMILES. The function that had been implemented previously did not enumerate all possible valid SMILES due to restrictions in some of the enumeration rules. 
Computational tasks were performed on NVIDIA cards. The code is available upon request in both Tensorflow (\citet{TF}) and Matlab (\citet{matlab}).
\par First we discuss how SMILES augmentation influences the CNF model and then we compare the best performances of the CNF model with existing models, such as deep neural networks (DNN) (\citet{wu18}), ConvGraph (\citet{tran16}) and TextCNN (\citet{kim14}) already implemented on OCHEM. When available, we use the benchmark results presented by \citet{wu18}.
\par The CNF model along with SMILES augmentation is performed on several data sets, including both regression and classification tasks. The data were obtained from open source data bases, which are all available on OCHEM. In the regression setting, predictions are made on targets such as the MP, melting point (\citet{tetko14}), the BP, boiling point (\citet{brandmaier12}), the BCF, bioconcentration factor (\citet{brandmaier12}), FreeSolv, free solvation (\citet{wu18}), LogS (\citet{delaney04}), Lipo, lipophilicity (\citet{huuskonen00}), BACE (\citet{wu18}), the DHFR, dihydrofolate reductase inhibition (\citet{sutherland04}) and the LEL, lowest effect level (\citet{novotarskyi16}). We choose the Root Mean Square Error (RMSE) as a measure of comparison.
\par For classification, we test the following: HIV (\citet{wu18}), AMES (\citet{sushko10}), BACE (\citet{wu18}), Clintox (\citet{wu18}), Tox21 (\citet{wu18}), the BBBP, blood-brain barrier penetration (\citet{wu18}), JAK3 (\citet{suzuki00}), BioDeg (\citet{vorberg14}) and RP AR (\citet{rybacka15}), and compare them using the AUC (Area Under the Curve) measure. These sets include small sizes (of order 1e2), medium sizes (of order 1e3) and larger sizes (of order 1e4), as shown in Table~\ref{res_reg_class}.
\par
\begin{figure}
    \centering
      \includegraphics[scale=0.2]{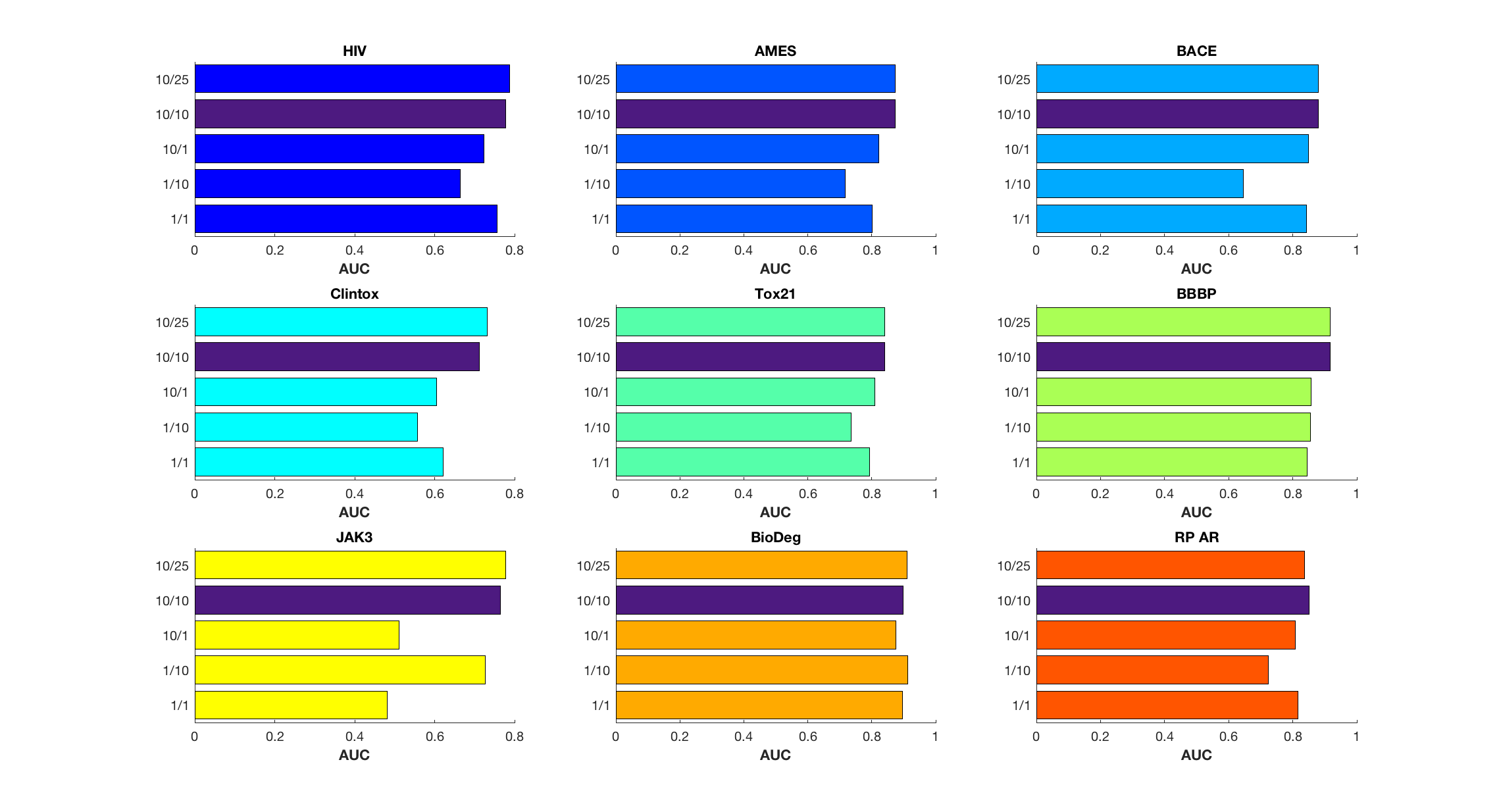}
    \caption{\small{The evaluation of the CNF model along with SMILES enumeration for classification tasks. Enumeration is done during training only (SMILES 10/1), testing only (SMILES (1/10), both (SMILES 10/10 and SMILES 10/25) or none (SMILES 1/1). The default parameter (SMILES 10/10) is displayed in purple. The higher the AUC value, the better the performance. The figure indicates that augmentation during training and testing (SMILES 10/10 or SMILES 10/25) certainly improves the accuracy of the CNF model.}}
    \label{fig:res_class_aug}
\end{figure}

\begin{figure}[]
    \centering
      \includegraphics[scale=0.3]{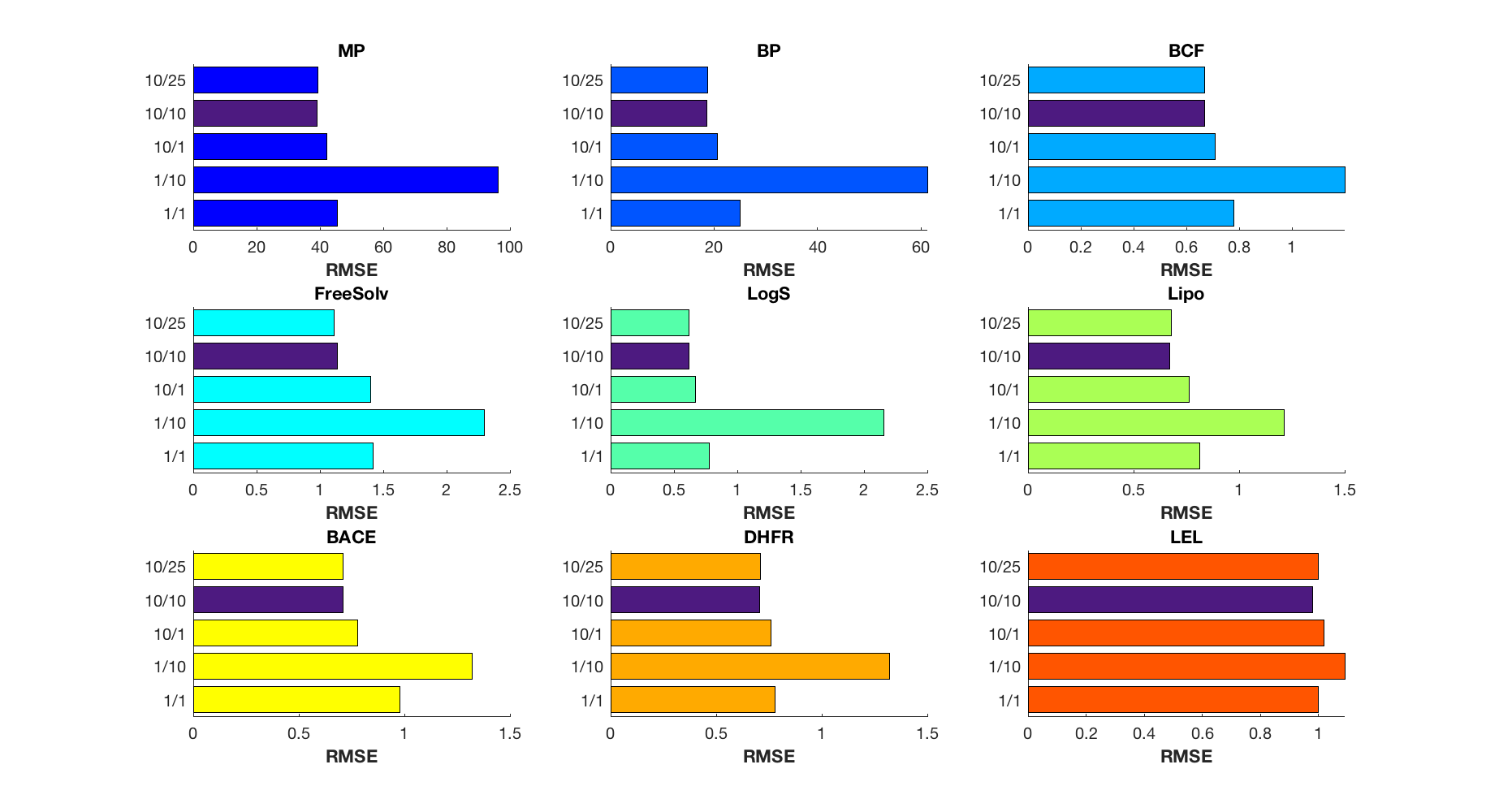}
    \caption{\small{The evaluation of the CNF model along with SMILES enumeration for regression tasks. Enumeration is done during training only (SMILES 10/1), testing only (SMILES (1/10), both (SMILES 10/10 and SMILES 10/25) or none (SMILES 1/1). The default parameter (SMILES 10/10) is displayed in purple. The lower the RMSE value, the better the performance. The figure indicates that augmentation during training and testing (SMILES 10/10 or SMILES 10/25) certainly improves the accuracy of the CNF model.}} 
    \label{fig:res_reg_aug}
\end{figure}

 In order to efficiently test the performances of the models, we use 5-fold cross-validation. The splitting of the data was done before the augmentation of SMILES. In order to test the effect of SMILES augmentation on the CNF model, we compare the following:
\begin{enumerate}
\item No augmentation on SMILES. We leave both train and test sets unchanged, for which we use the notation: SMILES 1/1. The considered SMILES are the canonical version.
\item SMILES augmentation during training only. If we generate $n$ random SMILES in the train set, where $n \in \mathbb{N}$, but leave the test set unchanged, we use the notation: SMILES n/1. The SMILES in the test set are the canonical versions.
\item SMILES augmentation during testing only. If we generate $m$ random SMILES in the test set, where $m \in \mathbb{N}$, but leave the train set unchanged, we note: SMILES 1/m. The SMILES in the train set are the canonical versions.
\item SMILES augmentation during both training and testing. If we generate $n$ random SMILES in the train set and $m$ random SMILES in the test set, where $n,m \in \mathbb{N}$, we note: SMILES n/m.
\end{enumerate}
The results of the simulations to test the effect of SMILES augmentation are detailed in Table~\ref{res_reg_class}. Based on the illustration of these results displayed in Figures~\ref{fig:res_class_aug} and \ref{fig:res_reg_aug}, we make the following observations. Augmenting during training (SMILES n/1) almost always improves predictive performances compared to leaving the data set untouched (SMILES 1/1). This observation confirms that SMILES augmentation during training can be interpreted as data augmentation, as discussed in Section~\ref{DA}. Augmenting during testing only (SMILES 1/m) almost always worsens the performances compared to leaving the data set untouched (SMILES 1/1). This shows that the CNF model is only able to map the SMILES to one instance of the enumeration and not to the original graph. In our implementation, the input SMILES are canonical and the CNF model received non-canonical SMILES for prediction. In this case, the neural network is confronted to new data, which it never had the opportunity to train on, leading to poor predictive power.
\par The situation where the best results are obtained is when both augmentation during training and testing is implemented. There is not only a data augmentation effect which acts on the train set but there is also an ensemble learning effect on the test set, as discussed in Section~\ref{DA}.
\par We choose the best models from Table~\ref{res_reg_class} and compare the results with the models that give the best results in Deepchem (\citet{wu18}). Please refer to Table~\ref{res_cnf_vs_deepchem} for the results. We notice that in most cases, the CNF model performs at least as well as the best Deepchem models.
\par
\begin{figure}[]
    \centering
      \includegraphics[scale=0.25]{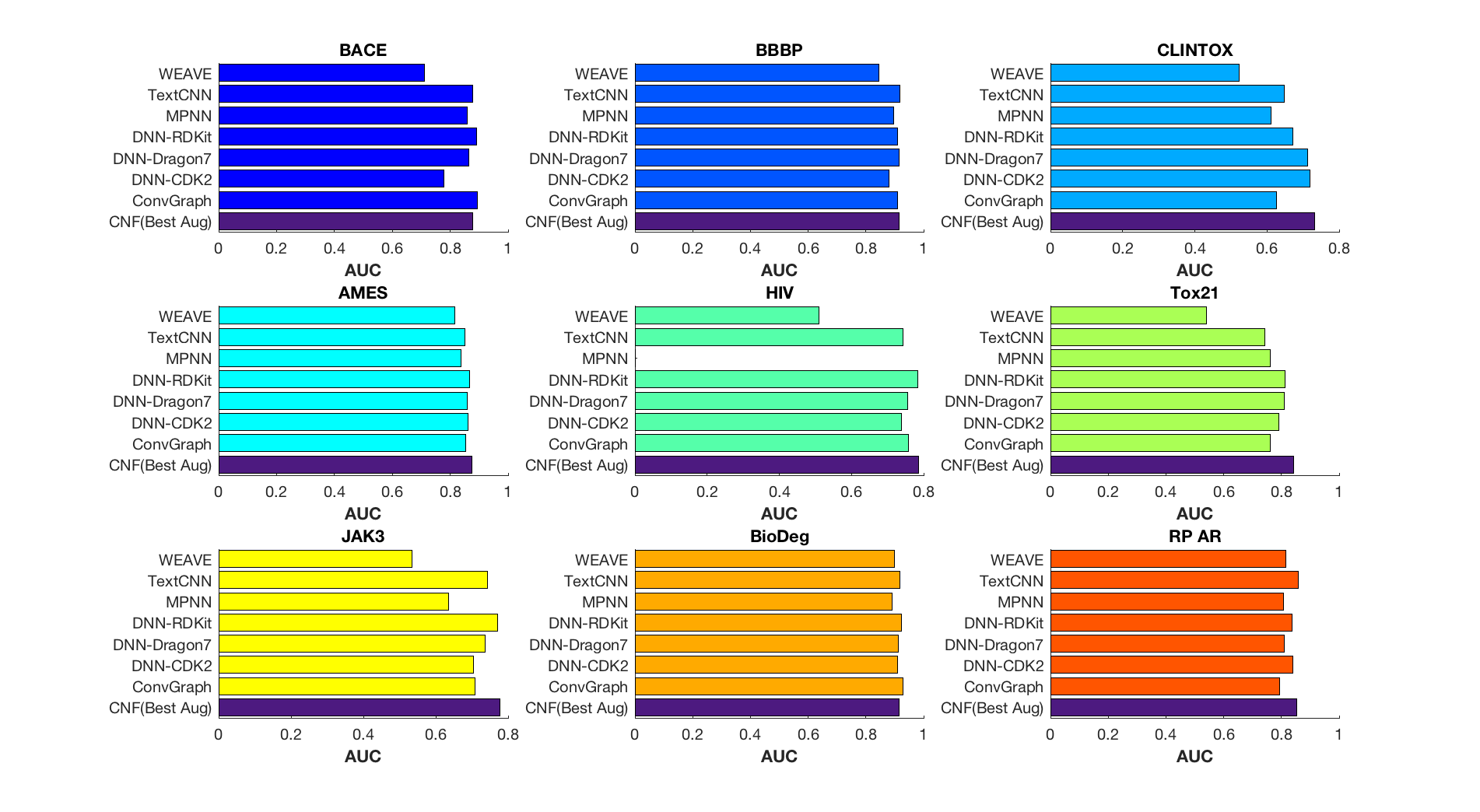}
    \caption{\small{Comparison of the best CNF model and other existing models for classification tasks. The best CNF model is displayed in purple. The higher the AUC value, the better the performance.  The figure indicates that the CNF model provides better accuracy compared to models built using the best descriptors and computational methods.}}
    \label{fig:res_class_mod}
\end{figure}

\begin{figure}[]
    \centering
      \includegraphics[scale=0.25]{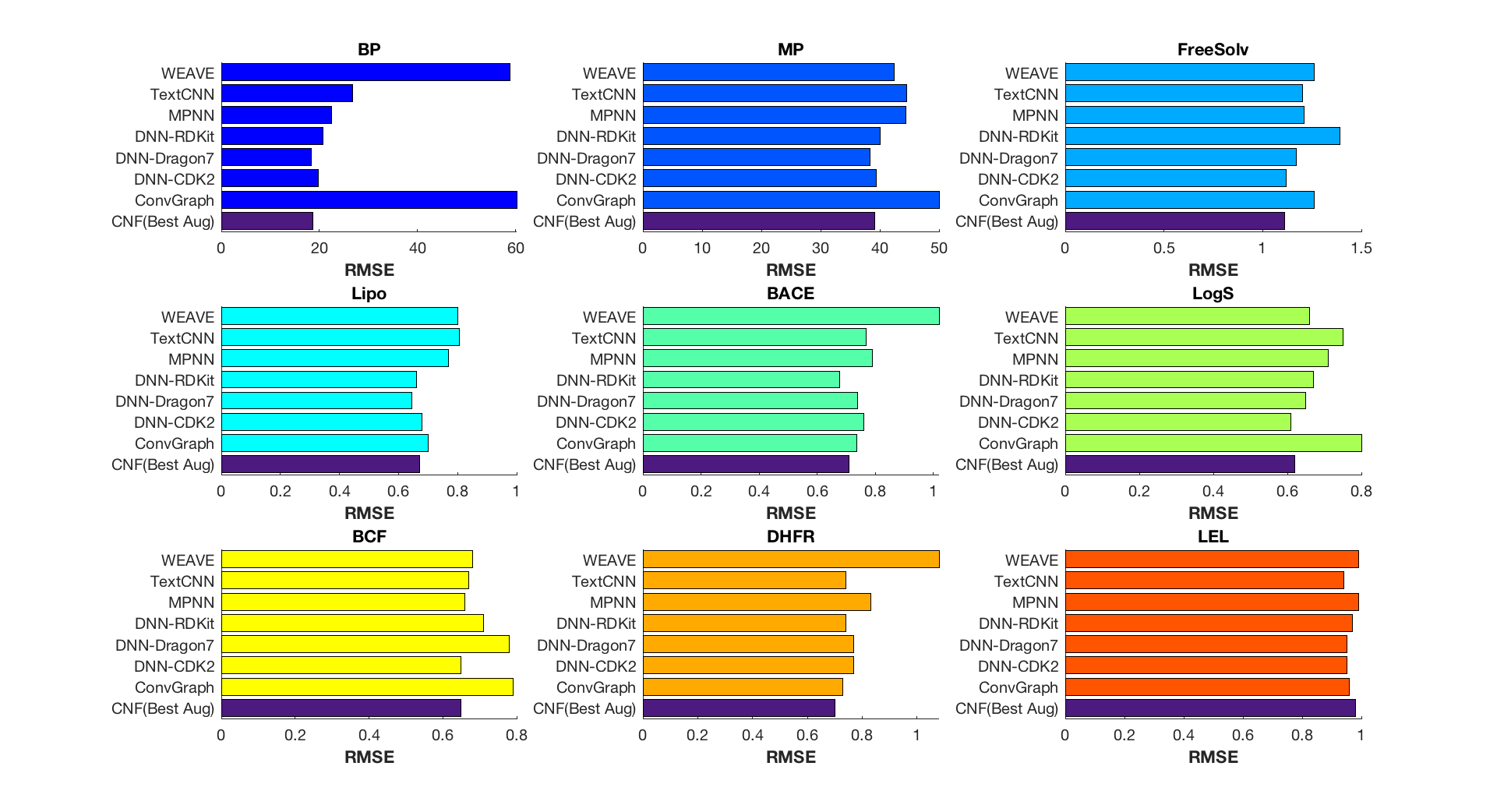}
    \caption{\small{Comparison of the best CNF model and other existing models for regression tasks. The best CNF model is displayed in purple. The lower the RMSE value, the better the performance. The figure indicates that the CNF model provides better accuracy compared to models built using the best descriptors and computational methods.}} 
    \label{fig:res_reg_mod}
\end{figure}

 Tables \ref{res_ochem_class} and \ref{res_ochem_reg} list the results of classification and regression tasks respectively using models available on OCHEM, where the simulations were done with the default parameters. Figures~\ref{fig:res_class_mod} and \ref{fig:res_reg_mod} give an illustration of these results. We notice that in most cases the CNF model provides better accuracy compared to models built using the best descriptors and computational methods.
\vspace{1.5cm}
\section{Conclusion and Future Work}
In this study, we apply machine learning algorithms on SMILES, which is a convenient linear graph notation containing information of a molecular structure. It is well known that there are many valid SMILES for a given molecule. The model we propose, namely the Convolutional Neural Fingerprint (CNF) model, seems to be able to extract the information about the local structure of a chemical compound thanks to convolutional layers. Using the multiplicity of SMILES as means of data augmentation also allows the CNF model to learn about the global structure. SMILES augmentation improves the CNF model predictions by approximately 15\% (and in the best of cases by approximately 25\%) which is enough to reach conventional human descriptor performances, such as the well known CDK2, Dragon or RDKit. These performances are observed on difficult target values such as the melting point of a molecular structure.
\par
Currently we have investigated molecular QSAR/QSPR (quantitative structure-activity/ property relationship) models. One of our objectives is to use the CNF model to also predict quantum chemistry atomic targets.
\par Another goal would be to apply augmentation not only on the prediction of molecular behaviour as we have tested but as well on reaction tasks.
\par The ensemble learning resulting from the augmentation during testing helps provide a metric to determine if a SMILES is in the applicability domain of the model. The mean and variance can provide information on the quality of the model. This approach is already implemented in OCHEM but has not been fully investigated.
\par Augmentation during training and testing is also under investigation for the TextCNN model.

\subsubsection*{Acknowledgements}
Talia Kimber would like to thank Eric Bruno, Guillaume Godin and Sebastian Engelke for their support and dedication in supervising this project which is part of her Master's Thesis.
\newline
Sebastian Engelke is grateful for financial support of the Swiss National Science Foundation.
\newline
Guillaume Godin thanks Greg Landrum, Esben J. Bjerrum and Takayuki Serizawa for the RDKit UGM Hackathon which took place in Cambridge in September 2018.
\newline
Guillaume Godin is grateful for the improvements made by Arvind Jayaraman on the MathWorks' deep learning toolbox.
\newline
The authors thank Alpha A. Lee as well as Philippe Schwaller for beneficial discussions.
\bibliographystyle{CUP}
\bibliography{bibliography}
\section{Supplementary Material}\label{supp_mat}
\begin{table}
\begin{center}
\begin{tabular}{cc}
\scalebox{0.7}{
\begin{tabular}{|| l | c | c | c ||}
\hline
Target & Size & Augmentation & RMSE\\
\hline
\hline
MP & 1\,9104 & 1/1 & 45.6 \\
 & & 10/1 & 42.8 \\
 & & 1/10 & 96.2 \\
 & & 10/10 & 39.2 \\
 & & \textbf{10/25} & \textbf{39.0} \\
\hline
BP & 1\,1893 & 1/1 & 25.0 \\
 & & 10/1 & 20.7 \\
 & & 1/10 & 61.2 \\
 & & 10/10 & 18.6 \\
 & & \textbf{10/25} & \textbf{18.6} \\
\hline
BCF & 378 & 1/ 1 & 0.78 \\
 & & 10/1 & 0.71\\
 & & 1/10 & 1.20 \\
 & & 10/10 & 0.65 \\
 & & \textbf{10/25} & \textbf{0.65} \\
\hline
FreeSolv & 642 & 1/ 1 & 1.42 \\
 & & 10/1 & 1.40 \\
 & & 1/10 & 2.30 \\
 & & 10/10 & 1.14\\
 & & \textbf{10/25} & \textbf{1.11}\\
\hline
LogS & 1\,311 & 1/ 1 & 0.78 \\
 & & 10/1 & 0.67 \\
 & & 1/10 & 2.16 \\
 & & 10/10 & 0.62 \\
 & & \textbf{10/25} & \textbf{0.62} \\
\hline
Lipo & 4\,200 & 1/ 1 & 0.81 \\
 & & 10/1 & 0.76 \\
 & & 1/10 & 1.21 \\
 & & \textbf{10/10} & \textbf{0.67} \\
 & & 10/25 & 0.68 \\
\hline
BACE & 1\,513 & 1/ 1 & 0.98 \\
 & & 10/1 & 0.78 \\
 & & 1/10 & 1.32 \\
 & & 10/10 & 0.71 \\
 & & \textbf{10/25} & \textbf{0.71}\\
 \hline
DHFR & 739 & 1/ 1 & 0.78 \\
 & & 10/1 & 0.76 \\
 & & 1/10 & 1.32 \\
 & & \textbf{10/10} & \textbf{0.70}\\
 & & 10/25 & 0.71\\
 \hline
LEL & 483 & 1/ 1 & 1.0 \\
 & & 10/1 & 1.0 \\
 & & 1/10 & 1.1 \\
 & & 10/10 & 1.0\\
 & & \textbf{10/25} & \textbf{1.0}\\
 \hline
\end{tabular}}
& 
\scalebox{0.7}{
\begin{tabular}{|| l | c | c | c ||}
\hline
Target & Size & Augmentation & AUC \\
\hline
\hline
HIV & 4\,1127 &1/1 & 0.76\\
 & & 10/1 & 0.72\\
 & & 1/10 & 0.67 \\
 & & 10/10 & 0.78 \\
 & & \textbf{10/25} & \textbf{0.79} \\
\hline
AMES & 6\,542 & 1/ 1 & 0.80 \\
 & & 10/ 1 & 0.82 \\
 & & 1/10 & 0.72 \\
 & & 10/10 & 0.87 \\
 & & \textbf{10/25} & \textbf{0.87} \\
\hline
BACE & 1\,513 & 1/ 1 & 0.84 \\
 & & 10/1 & 0.85 \\
 & & 1/10 & 0.65 \\
 & & 10/10 & 0.88 \\
 & & \textbf{10/25} & \textbf{0.88} \\
\hline
Clintox & 1\,478 & 1/ 1 & 0.62 \\
 & & 10/1 & 0.61 \\
 & & 1/10 & 0.56 \\
 & & 10/10 & 0.71 \\
 & & \textbf{10/25} & \textbf{0.73} \\
\hline
Tox21 & 7\,831 & 1/ 1 & 0.79 \\
 & & 10/1 & 0.81 \\
 & & 1/10 & 0.74 \\
 & & 10/10 & 0.84 \\
 & & \textbf{10/25} & \textbf{0.84} \\
\hline
BBBP & 2\,039 & 1/ 1 & 0.84 \\
 & & 10/1 & 0.86 \\
 & & 1/10 & 0.86 \\
 & & 10/10 & 0.92 \\
 & & \textbf{10/25} & \textbf{0.92} \\ 
 \hline
JAK3 & 886 & 1/ 1 & 0.48 \\
 & & 10/1 & 0.51 \\
 & & 1/10 & 0.73 \\
 & & 10/10 & 0.76 \\
 & & \textbf{10/25} & \textbf{0.78} \\ 
 \hline
BioDeg & 1\,737 & 1/ 1 & 0.90 \\
 & & 10/1 & 0.88 \\
 & & 1/10 & 0.91 \\
 & & 10/10 & 0.90 \\
 & & \textbf{10/25} & \textbf{0.91} \\ 
 \hline
RP AR & 930 & 1/ 1 & 0.82 \\
 & & 10/1 & 0.81 \\
 & & 1/10 & 0.72 \\
 & & \textbf{10/10} & \textbf{0.85} \\ 
 & & 10/25 & 0.83 \\
 \hline
\end{tabular}}
\end{tabular}
\caption{Summary of the results of the simulations using the CNF model with the multiplicity of SMILES. The first column represents the target value (the left and right tables consider regression and classification tasks, respectively). The second column represents the size of the initial data set. The third column shows the type of augmentation that is performed on the data set, and the last column is the measure chosen to compare the performance of the CNF model, namely the RMSE and the AUC in regression and classification tasks, respectively. The augmentation which gives the best result is displayed in bold, indicating that augmentation during training and testing (SMILES 10/10 or SMILES 10/25) certainly improves the accuracy of the model.}
\label{res_reg_class}
\end{center}
\end{table}
\begin{table}
\begin{center}
\begin{minipage}{0.7\textwidth}
\begin{center}
\begin{tabular}{c c}
\scalebox{0.7}{\begin{tabular}{| l | l c |}
\hline
Target & Model & Performance (RMSE)\\
\hline
\hline
MP & \textbf{CNF} & \textbf{39.2}\\
& Deepchem & NA\footnote{Results for these targets are not available (NA) in the original paper by \citet{wu18}.} \\
\hline
BP & \textbf{CNF} & \textbf{18.6}\\
& Deepchem & NA \\
\hline
BCF & \textbf{CNF} & \textbf{0.65}\\
& Deepchem & NA \\
\hline
FreeSolv & \textbf{CNF} & \textbf{1.11}\\
& MPNN & 1.15 \\
\hline
LogS & \textbf{CNF} & \textbf{0.62}\\
& Deepchem & NA \\
\hline
Lipo & CNF & 0.671\\
& \textbf{ConvGraph} & \textbf{0.66} \\
\hline
BACE & \textbf{CNF} & \textbf{0.71} \\
& Deepchem & NA \\
\hline
DHFR & \textbf{CNF} & \textbf{0.7} \\
& Deepchem & NA \\
\hline
LEL & \textbf{CNF} & \textbf{1} \\
& Deepchem & NA \\
\hline
\end{tabular}}
&
\scalebox{0.7}{\begin{tabular}{| l | l c |}
\hline
Target & Model & Performance (AUC)\\
\hline
\hline
HIV & CNF & 0.79\\
& \textbf{KernelSVM} & \textbf{0.792} \\
\hline
AMES & \textbf{CNF} & \textbf{0.87}\\
& Deepchem & NA \\
\hline
BACE & \textbf{CNF} & \textbf{0.88}\\
& RF & 0.867 \\
\hline
Clintox & CNF & 0.73\\
& \textbf{Weave} & \textbf{0.832} \\
\hline
Tox21 & \textbf{CNF} & \textbf{0.84} \\
& ConvGraph & 0.829 \\
\hline
BBBP & \textbf{CNF} & \textbf{0.92}\\
& KernelSVM & 0.729 \\
\hline
JAK3 & \textbf{CNF} & \textbf{0.78}\\
& Deepchem & NA \\
\hline
BioDeg & \textbf{CNF} & \textbf{0.92}\\
& Deepchem & NA \\
\hline
RP AR & \textbf{CNF} & \textbf{0.85}\\
& Deepchem & NA \\
\hline
\end{tabular}}
\end{tabular}
\end{center}
\end{minipage}
\end{center}
\caption{Comparison of the best CNF model with best results in Deepchem (\citet{wu18}), using regression tasks (left table) and classification tasks (right table). The first column represents the target value. The second column lists the models. The third column shows the performance of each model. The best model is displayed in bold, indicating that the best CNF model usually performs better than the best Deepchem model.}
\label{res_cnf_vs_deepchem}
\end{table}
\begin{table}
\begin{center}
\begin{minipage}{1\textwidth}
\begin{center}
\begin{tabular}{ccc}
\scalebox{0.7}{
\begin{tabular}{| l | l c |}
\hline
Target & Model (5 CV) & AUC\\
\hline
\hline
HIV & \textbf{CNF} & \textbf{0.79}\\
& ConvGraph & 0.76 \\
& TextCNN & 0.74 \\
& MPNN & NA \footnote{Result for this target is not available (NA).}\\
& Weave & 0.51 \\
\hdashline
& DNN (Dragon7) & 0.76 \\
& DNN (CDK2)& 0.74 \\
& \textbf{DNN (RDKit)}& \textbf{0.78} \\
& DNN (PyDescriptor)& 0.77 \\
\hline
AMES & \textbf{CNF} & \textbf{0.87}\\
& ConvGraph & 0.85 \\
& TextCNN & 0.85 \\
& MPNN & 0.84 \\
& Weave & 0.82 \\
\hdashline
& DNN (Dragon7)& 0.86 \\
& DNN (CDK2)& 0.86 \\
& \textbf{DNN (RDKit)} & \textbf{0.87} \\
& DNN (PyDescriptor)& 0.86 \\
\hline
BACE & \textbf{CNF} & \textbf{0.88}\\
& \textbf{ConvGraph} & \textbf{0.89} \\
& TextCNN & 0.88 \\
& MPNN & 0.86 \\
& Weave & 0.71 \\
\hdashline
& DNN (Dragon7)& 0.87 \\
& DNN (CDK2)& 0.78 \\
& \textbf{DNN (RDKit)} & \textbf{0.89} \\
& DNN (PyDescriptor)& 0.86 \\
\hline
\end{tabular}}
& 
\scalebox{0.7}{
\begin{tabular}{| l | l c |}
\hline
Target & Model (5 CV) & AUC\\
\hline
\hline
Clintox & \textbf{CNF} & \textbf{0.73}\\
& ConvGraph & 0.63 \\
& TextCNN & 0.65 \\
& MPNN & 0.61 \\
& Weave & 0.52 \\
\hdashline
& \textbf{DNN (Dragon7)}& \textbf{0.71} \\
& DNN (CDK2)& 0.72 \\
& DNN (RDKit)& 0.67 \\
& DNN (PyDescriptor)& 0.71 \\
\hline
Tox21 & \textbf{CNF} & \textbf{0.84} \\
& ConvGraph & 0.76 \\
& TextCNN & 0.74 \\
& MPNN & 0.76 \\
& Weave & 0.54 \\
\hdashline
& \textbf{DNN (Dragon7)}& \textbf{0.81} \\
& DNN (CDK2)& 0.79 \\
& DNN (RDKit)& 0.81 \\
& DNN (PyDescriptor)& 0.81 \\
\hline
BBBP & \textbf{CNF} & \textbf{0.92}\\
& ConvGraph & 0.91 \\
& TextCNN & 0.92 \\
& MPNN & 0.90 \\
& Weave & 0.84 \\
\hdashline
& \textbf{DNN (Dragon7)} & \textbf{0.92} \\
& DNN (CDK2)& 0.88 \\
& DNN (RDKit)& 0.91 \\
& DNN (PyDescriptor)& 0.91 \\
\hline
\end{tabular}}
&
\scalebox{0.7}{
\begin{tabular}{| l | l c |}
\hline
Target & Model (5 CV) & AUC\\
\hline
\hline
JAK3 & \textbf{CNF} & \textbf{0.78} \\
& ConvGraph & 0.71 \\
& TextCNN & 0.74 \\
& MPNN & 0.64 \\
& Weave & 0.53 \\
\hdashline
& DNN (Dragon7)& 0.74 \\
& DNN (CDK2)& 0.71 \\
& \textbf{DNN (RDKit)}& \textbf{0.77} \\
& DNN (PyDescriptor)& 0.76 \\
\hline
BioDeg & \textbf{CNF} & \textbf{0.92} \\
& \textbf{ConvGraph} & \textbf{0.93} \\
& TextCNN & 0.92 \\
& MPNN & 0.89 \\
& Weave & 0.90 \\
\hdashline
& DNN (Dragon7)& 0.91 \\
& DNN (CDK2)& 0.91 \\
& \textbf{DNN (RDKit)}& \textbf{0.92} \\
& DNN (PyDescriptor)& 0.91 \\
\hline
RP AR & \textbf{CNF} & \textbf{0.85}\\
& ConvGraph & 0.80 \\
& TextCNN & 0.85 \\
& MPNN & 0.81 \\
& Weave & 0.82 \\
\hdashline
& DNN (Dragon7)& 0.81 \\
& \textbf{DNN (CDK2)}& \textbf{0.84} \\
& DNN (RDKit)& 0.84 \\
& DNN (PyDescriptor)& 0.83 \\
\hline
\end{tabular}}
\end{tabular}
\end{center}
\end{minipage}
\end{center}
\caption{Comparison of the best CNF model with results on OCHEM, using 5-fold cross-validation on 9 classification tasks; HIV, AMES and BACE in the first table, Clintox, Tox21 and BBBP in the second table and JAK3, BioDeg and RP AR in the third table. The first column represents the target value. The second column lists the models. The third column shows the AUC, Area Under the Curve, value. The top-performing models which are shown in bold clearly demonstrate that the CNF model provides better accuracy compared to models built using the best descriptors and computational methods.}
\label{res_ochem_class}
\end{table}
\begin{table}
\begin{center}
\begin{minipage}{1\textwidth}
\begin{center}
\begin{tabular}{ccc}
\scalebox{0.7}{
\begin{tabular}{| l | l c |}
\hline
Target & Model (5 CV) & RMSE\\
\hline
\hline
MP & \textbf{CNF} & \textbf{39.2}\\
& ConvGraph & 50.0 \\
& TextCNN & 44.5 \\
& MPNN & 44.4 \\
& Weave & 42.4 \\
\hdashline
& \textbf{DNN (Dragon7)} & \textbf{38.4} \\
& DNN (CDK2)& 39.4 \\
& DNN (RDKit)& 40.0 \\
& DNN (PyDescriptor)& 41.0 \\
\hline
BP & \textbf{CNF} & \textbf{18.6}\\
& ConvGraph & 60.2 \\
& TextCNN & 26.8 \\
& MPNN & 22.4 \\
& Weave & 58.8 \\
\hdashline
& DNN (Dragon7)& 18.3 \\
& DNN (CDK2)& 19.8 \\
& DNN (RDKit)& 20.8 \\
& \textbf{DNN (PyDescriptor)}& \textbf{18.1} \\
\hline
BCF & \textbf{CNF} & \textbf{0.65}\\
& ConvGraph & 0.79 \\
& TextCNN & 0.68 \\
& MPNN & 0.66 \\
& Weave & 0.68 \\
\hdashline
& DNN (Dragon7)& 0.78 \\
& \textbf{DNN (CDK2)} & \textbf{0.65} \\
& DNN (RDKit)& 0.71 \\
& DNN (PyDescriptor)& 0.66 \\
\hline
\end{tabular}}
& 
\scalebox{0.7}{
\begin{tabular}{| l | l c |}
\hline
Target & Model (5 CV) & RMSE\\
\hline
\hline
FreeSolv & \textbf{CNF} & \textbf{1.11}\\
& ConvGraph & 1.26 \\
& TextCNN & 1.20 \\
& MPNN & 1.21 \\
& Weave & 1.26 \\
\hdashline
& DNN (Dragon7)& 1.17 \\
& DNN (CDK2)& 1.12 \\
& DNN (RDKit)& 1.39 \\
& \textbf{DNN (PyDescriptor)}& \textbf{1.09} \\
\hline
LogS & \textbf{CNF} & \textbf{0.62}\\
& ConvGraph & 0.80 \\
& TextCNN & 0.74 \\
& MPNN & 0.71 \\
& Weave & 0.66 \\
\hdashline
& DNN (Dragon7)& 0.65 \\
& \textbf{DNN (CDK2)}& \textbf{0.61} \\
& DNN (RDKit)& 0.67 \\
& DNN (PyDescriptor)& 0.63 \\
\hline
Lipo & \textbf{CNF} & \textbf{0.67}\\
& ConvGraph & 0.70 \\
& TextCNN & 0.81 \\
& MPNN & 0.77 \\
& Weave & 0.80 \\
\hdashline
& \textbf{DNN (Dragon7)}& \textbf{0.65} \\
& DNN (CDK2)& 0.68 \\
& DNN (RDKit)& 0.66 \\
& DNN (PyDescriptor) & 0.65 \\
\hline
\end{tabular}}
&
\scalebox{0.7}{
\begin{tabular}{| l | l c |}
\hline
Target & Model (5 CV) & RMSE\\
\hline
\hline
BACE & \textbf{CNF} & \textbf{0.71} \\
& ConvGraph & 0.74 \\
& TextCNN & 0.77 \\
& MPNN & 0.79 \\
& Weave & 1.02 \\
\hdashline
& DNN (Dragon7)& 0.74 \\
& DNN (CDK2)& 0.76 \\
& \textbf{DNN (RDKit)} & \textbf{0.68} \\
& DNN (PyDescriptor)& 0.72 \\
\hline
DHFR & \textbf{CNF} & \textbf{0.70} \\
& ConvGraph & 0.73 \\
& TextCNN & 0.74 \\
& MPNN & 0.83 \\ 
& Weave & 1.08 \\
\hdashline
& DNN (Dragon7)& 0.77 \\
& DNN (CDK2)& 0.77 \\
& \textbf{DNN (RDKit)}& \textbf{0.74} \\
& DNN (PyDescriptor)& 0.76 \\
\hline
LEL & \textbf{CNF} & \textbf{0.98}\\
& \textbf{ConvGraph} & \textbf{0.96} \\
& \textbf{TextCNN} & \textbf{0.94} \\
& MPNN & 0.99 \\
& Weave & 0.99 \\
\hdashline
& \textbf{DNN (Dragon7)}& \textbf{0.95} \\
& DNN (CDK2)& 0.95 \\
& DNN (RDKit)& 0.97 \\
& DNN (PyDescriptor)& 1.02 \\
\hline
\end{tabular}}
\end{tabular}
\end{center}
\end{minipage}
\end{center}
\caption{Comparison of the best CNF model with results in OCHEM, using 5-fold cross-validation on 9 regression tasks; MP, BP and BCF in the first table, FreeSolv, LogS and Lipo in the second table and BACE, DHFR and LEL in the third table. The first column represents the target value. The second column lists the models. The third column shows the RSME, Root Mean Square Error, value. The top-performing models which are shown in bold clearly demonstrate that the CNF model provides better accuracy compared to models built using the best descriptors and computational methods.}
\label{res_ochem_reg}
\end{table}
Let $S_1, \dots, S_N$ be $N$ initial SMILES of a given data set and suppose that for each SMILES, we generate $n$ random SMILES, namely
$$
S_1^1,\dots, S_1^n, S_2^1, … S_2^n,\dots, S_N^1, \dots, S_N^n.
$$
Suppose we consider the CNF model with only one convolutional layer. For $i = 1,\dots, N$ and $j=1,\dots, n$, let $\NFP _i^j$ be the neural fingerprint associated with the $j^{th}$ generated SMILES of the $i^{th}$ initial SMILES. Then we expect that the CNF model learns that $\forall i, l \in \{ 1,\dots, N \} ,\, \forall j, k, m \in \{1,\dots n\}$
$$
||\NFP _i^j - \NFP _i^k ||_2^2 \leq || \NFP _i^j - \NFP _l^m||_2^2,
$$
that is the Euclidian distance between neural fingerprints generated by the same SMILES is smaller or equal than the Euclidian distance between neural fingerprints generated by two different SMILES.
\par
\begin{figure*}
    \centering
    \begin{subfigure}[b]{0.7\textwidth}
      \centering
      \includegraphics[width=\textwidth]{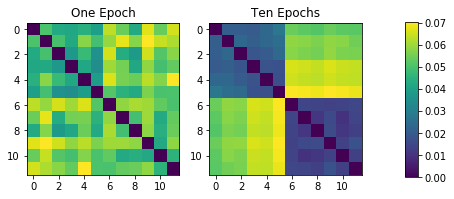}
      \caption[Network2]%
      {{\small Close up of the distance matrix $D$ after one and ten epochs respectively of the CNF model, namely $D_1$ and $D_{10}$. We notice the creation of two main clusters after ten epochs.}}  
      \label{fig:4000}
    \end{subfigure}
    \\
    \begin{subfigure}[b]{0.7\textwidth} 
      \centering 
      \includegraphics[width=\textwidth]{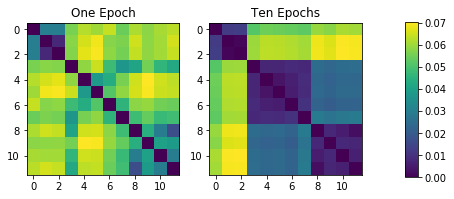}
      \caption[]%
      {{\small Different close up of the distance matrix $D$ after one and ten epochs respectively of the CNF model, namely $D_1$ and $D_{10}$. We notice the creation of three main clusters after ten epochs.}}  
      \label{fig:8000}
    \end{subfigure}
    \caption
    {Suppose we generate $n$ random SMILES from a data set of size $N$. For $i = 1,\dots, N$ and $j=1,\dots, n$, let $\NFP _i^j$ be the neural fingerprint associated with the $j^{th}$ generated SMILES of the $i^{th}$ initial SMILES, obtained with the CNF model. Let $D_1, D_{10}$ be the Euclidian distance matrices associated with all NFPs after one and ten epochs, respectively. Figure~\ref{fig:4000} and \ref{fig:8000} show a close up of $D_1$ and $D_{10}$. We notice that as the CNF model trains, the Euclidian distance between neural fingerprints generated by the same SMILES is smaller than the Euclidian distance between neural fingerprints generated by two different SMILES. } 
    \label{dist_mat_close_up}
\end{figure*}

 The following results give us reason to believe that the statement above is true.
Let $D_1$ and $D_{10} \in \mathbb{R}^{Nn \times Nn}$ be the distance matrices obtained after one and ten epochs respectively (see Figure~\ref{dist_mat_close_up}). We notice that as the CNF model trains, the Euclidian distance between neural fingerprints generated by the same SMILES is smaller than the Euclidian distance between neural fingerprints generated by two different SMILES. 
%
%
\end{document}